\documentclass[11pt]{article}
\usepackage[utf8]{inputenc} 
\usepackage[T1]{fontenc}    
\usepackage{hyperref}       
\usepackage{url}            
\usepackage{booktabs}       
\usepackage{nicefrac}       
\usepackage{microtype}      
\usepackage{xcolor}         
\usepackage[textsize = scriptsize]{todonotes}
\usepackage{amsmath,amssymb,amsfonts}
\usepackage{amsthm}
\usepackage{mathtools}
\mathtoolsset{showonlyrefs=false}
\usepackage{bm}
\usepackage{subcaption}
\usepackage{algorithm}
\usepackage{bbm}
\usepackage{algpseudocode}
\usepackage{comment}
\usepackage{tikz}
\usepackage{enumerate}
\usepackage{authblk}
\usepackage{tikz}
\usepackage{pgfplots}
\usepackage[export]{adjustbox}
\usetikzlibrary{spy}
\usepackage{geometry}

\theoremstyle{definition}

\begin{document}

\title{PnP-DA: Towards Principled Plug-and-Play Integration of Variational Data Assimilation and Generative Models}

\author[1,2,4]{Yongquan Qu}
\author[2,3]{Matthieu Blanke}
\author[2,3]{Sara Shamekh}
\author[1,2]{Pierre Gentine}

\affil[1]{Department of Earth and Environmental Engineering, Columbia University, New York, USA}
\affil[2]{NSF Center for Learning the Earth with Artificial Intelligence and Physics (LEAP), New York, USA}
\affil[3]{Courant Institute for Mathematical Science, New York University, New York, USA}
\affil[4]{Pasteur Labs, New York, USA}

\date{}  

\maketitle

\begin{abstract}
Earth system modeling presents a fundamental challenge in scientific computing: capturing complex, multiscale nonlinear dynamics in computationally efficient models while minimizing forecast errors caused by necessary simplifications. Even the most powerful AI- or physics-based forecast system suffer from gradual error accumulation. Data assimilation (DA) aims to mitigate these errors by optimally blending (noisy) observations with prior model forecasts, but conventional variational methods often assume Gaussian error statistics that fail to capture the true, non-Gaussian behavior of chaotic dynamical systems. We propose \textsc{PnP‑DA}, a Plug‑and‑Play algorithm that alternates (1) a lightweight, gradient‑based analysis update—using a Mahalanobis‐distance misfit on new observations—with (2) a single forward pass through a pretrained generative prior conditioned on the background forecast via a conditional Wasserstein coupling. This strategy relaxes restrictive statistical assumptions and leverages rich historical data without requiring an explicit regularization functional, and it also avoids the need to backpropagate gradients through the complex neural network that encodes the prior during assimilation cycles. Experiments on standard chaotic testbeds demonstrate that this strategy consistently reduces forecast errors across a range of observation sparsities and noise levels, outperforming classical variational methods. 
\end{abstract}


\section{Introduction}

\begin{figure}[t]
\centering
\centerline{\includegraphics[width=\columnwidth]{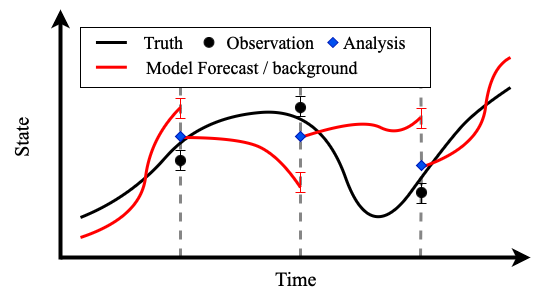}}
\caption{Detailed sketch of data assimilation.}
\label{fig:sketch}
\end{figure}

Earth system modeling presents one of the most computationally demanding challenges in scientific computing, requiring simulation of complex physical processes across multiple spatial and temporal scales \cite{palmer2019scientific}. A fundamental limitation in this field is that computational constraints and incomplete physical understanding necessitate the use of simplified physical models that cannot fully resolve all scales. This scale disparity introduces systematic errors that propagate through the forecasts and diminish the prediction accuracy \cite{palmer2001nonlinear}. Data assimilation (DA) forms the foundation of modern weather forecasting and Earth system prediction, addressing the inverse problem of estimating the current state from sparse, noisy observations and prior prediction (called background) coming from the imperfect model \cite{carrassi2018data, cheng2023machine} for a better initial condition (called analysis) of subsequent forecast (see Figure \ref{fig:sketch}). It can also be used to perform parameter estimation to mitigate model misspecifications \cite{evensen2009ensemble, qu2024joint}.

While machine learning (ML) approaches have begun transforming the forecasting component of weather prediction \cite{bi2023accurate, price2025probabilistic}, their application to the underlying DA problem remains an active research area with substantial untapped potential \cite{xu2025fuxi}. Recent work has advanced the integration of DA with modern ML tools through differentiable numerical solvers or differentiable surrogates that facilitate gradient computation within variational DA frameworks \cite{farchi2023online, qu2023can, qu2024joint, solvik20244d, xiao2024vae, fan2025physically}. Other researchers have addressed the probabilistic formulation of DA using modern generative models, for example, diffusion model \cite{huang2024diffda, finn2024generative, qu2024deep, manshausen2024generative, blanke2025strictly}, while some have focused on developing novel filtering and optimization approaches with ML \cite{bao2024ensemble, blanke2024neural, pasmans2025ensemble}.

Traditional variational DA methods assume Gaussian error statistics for model errors. This assumption makes the assimilation problem tractable but blind to the non‑Gaussian, state‑dependent errors in chaotic flows. Some state-of-the-art strategies use ML for variational DA to address this limitation through two primary approaches: mapping background errors into a near Gaussian latent space via variational autoencoders \cite{xiao2024vae} or mapping the physical state(s) into a latent space with near diagonal covariance structure \cite{fan2025physically}, both leveraging autoencoding procedures. While these methods have demonstrated great empirical performance with ML surrogates for atmospheric modeling, each has limitations. The former faces challenges because the change of measure introduces a costly Jacobian log-determinant which is both expensive to compute and numerically unstable in arbitrary deep networks. As a simplification, the Jacobian log-determinant term is therefore omitted for numerical stability in \cite{xiao2024vae}. Consequently, high-density regions of the true posterior—where the decoder locally compresses latent volume—are no longer preferentially weighted. The resulting MAP estimate still fits the observations but may systematically mislocate modes relative to the correct Bayesian push-forward, and no longer corresponds to a true maximum-a-posteriori estimate under the intended generative model. By performing variational DA directly in a latent space representing the physical states—rather than transforming the error statistics—the approach avoids the need to compute an intractable Jacobian log-determinant; however, projection back to the physical space inevitably introduces reconstruction error arising from the reduced-order representation. Both of these algorithms require additional gradient backpropagation through the highly nonlinear neural network during optimization iterations.

Building on this review of ML-enhanced variational DA, our approach departs fundamentally from latent-space autoencoding strategies. We aim to develop a more powerful background-state prior that can fully leverage the extensive record of historical forecasts and reanalysis data. We also seek to keep the analysis step numerically light by avoiding intractable Jacobian corrections or additional gradient backpropagation through learned autoencoders. Furthermore, instead of training models to simply match reanalysis data in an end-to-end fashion, our goal is to optimize them to perform well on future, unseen observations.

The Plug-and-Play (PnP) framework offers an established approach for solving inverse problems by integrating powerful learned priors directly into the optimization process, without requiring an explicit regularization functional \cite{venkatakrishnan2013plug}. Recent advances in generative modeling have introduced expressive priors that naturally fit within this framework \cite{zhu2023denoising, wu2024principled, martin2024pnp}. Data assimilation (DA) is fundamentally an ill-posed inverse problem, where the goal is to infer the true system state from incomplete and noisy observations. Variational DA addresses this by imposing a Gaussian prior centered on a background forecast to regularize the solution. In contrast, PnP methods with generative priors offer a more flexible alternative: instead of relying on simplified error statistics, they leverage rich, data-driven priors learned from historical data to constrain the solution space, improving robustness and realism in ill-posed settings. However, while these methods show promise for static inverse problems like image reconstruction, DA presents unique challenges. The solution to the inverse problem becomes an initial condition for a potentially chaotic dynamical system simulation, where visual quality doesn't guarantee prediction accuracy or even numerical stability. More importantly, unlike traditional inverse problems, where a known degradation operator maps the true signal to the observed one, there is no explicit or well-defined mapping from the true physical state to the background forecast in data assimilation. This complicates the direct application of standard PnP methods, which typically rely on such forward models to guide optimization.

In this paper, we propose a novel data-assimilation framework based on a Plug-and-Play algorithm that combines a variational DA update on incoming observations with a flow-matching prior on the background forecast, drawing inspiration from PnP-Flow \cite{martin2024pnp}. Unlike the standard optimal transport (OT) flow matching used in PnP-Flow, we use the OT Bayesian flow matching \cite{chemseddine2024conditional} as the generative prior. This approach conditions the flow on the background forecast field by incorporating a conditional Wasserstein cost into the optimal transport coupling. This cost strongly penalizes deviations from the background, ensuring that the learned velocity fields serve as a Bayesian-optimal correction—specifically, they minimize the expected squared transport cost given the background field. At each analysis step, we perform a lightweight gradient descent to reduce the observation misfit, followed by the pre-trained flow-based denoising operator. We model the observation misfit as a Mahalanobis distance, which is algebraically equivalent to the negative Gaussian log-likelihood up to an additive constant but makes no explicit assumption of normality of the observation error. This method unites the variational assimilation framework--- which leverages observation gradient--- with the expressive power of flow-based models to capture complex conditional distributions, while also retaining interpretability due to the tractable properties of straight-line flows. This method offers a principled, flexible, and numerically stable assimilation framework that improved accuracy across different levels of observational sparsity and noise, as demonstrated in our preliminary cyclic DA experiments on 3 chaotic dynamical system testbeds.

\section{Methods}
\subsection{Problem Formulation}
DA can be framed as a specific but extended case of general “static’’ inverse problems where we aim to recover the true physical state \(x_t \in \mathbb{R}^d\) at time \(t\) from noisy, incomplete observations \(y_t \in \mathbb{R}^m\):

\begin{equation}
  y_t = \mathcal{H}_t(x_t) + \xi_t,
  \label{eq:ObservationModel}
\end{equation}
where \(\mathcal{H}_t : \mathbb{R}^d \to \mathbb{R}^m\) is the (possibly nonlinear) observation operator at time \(t\) with \(m<d\), and \(\xi_t \in \mathbb{R}^m\) is the additive observation error. From a Bayesian perspective, the maximum a posteriori (MAP) estimate is the solution to the regularized optimization problem

\begin{equation}
  x_t^a = \arg\min_{x_t \in \mathbb{R}^d}
    \bigl\{\,F(x_t) + R(x_t)\bigr\},
  \label{eq:GeneralInverseProblem}
\end{equation}
where
\[
  F(x_t) := -\log p\bigl(y_t \mid x_t\bigr)
  \quad\text{and}\quad
  R(x_t) := -\log p(x_t)
\]
are the data‐fidelity and prior terms, respectively, and the solution \(x_t^a\) is called the analysis at time \(t\).
What distinguishes DA from static inverse problems is the presence of a dynamical model for the state, used as prior for the inverse problem:
\begin{equation}
  x_t = \mathcal{M}_t(x_{t-1}) + \eta_t,
  \label{eq:DynamicalModel}
\end{equation}
where \(\mathcal{M}_t : \mathbb{R}^d \to \mathbb{R}^d\) is an (imperfect) forecast model and \(\eta_t \in \mathbb{R}^d\) is model error. The analysis from \eqref{eq:GeneralInverseProblem} then provides an improved initial condition for the next forecast step. The background  \(x_t^b\) is defined as the system state obtained by propagating the previous analysis forward with the imperfect dynamical model, which suggests a natural prior: the true state \(x_t\) should lie near the background state. If we assume Gaussian distributions for the model and observation errors
\[
  x_t \sim \mathcal{N}(x_t^b,\,\textbf{B}_t), 
  \quad
  \xi_t \sim \mathcal{N}(0,\,\textbf{P}_t),
\]
then the objective becomes the standard 3D‐Var minimization function
\begin{equation}
  J(x_t)
  = \frac{1}{2} \bigl\|y_t - \mathcal{H}_t(x_t)\bigr\|^2_{\textbf{P}_t^{-1}}
  + \frac{1}{2} \bigl\|x_t - x_t^b\bigr\|^2_{\textbf{B}_t^{-1}}.
\end{equation}
When the observation operator is linear, denoted by \(\mathbf{H}\), and omitting subscript \(t\) for simplicity, the analytical solution is:
\begin{equation}
x^a =  {x}^b + \mathbf{B}\,\mathbf{H}^T
\bigl(\mathbf{H}\,\mathbf{B}\,\mathbf{H}^T + \mathbf{P}\bigr)^{-1}
\bigl(y - \mathbf{H}\,\mathbf{x}^b\bigr).
\label{eq:3dvar-sol}
\end{equation}
The same reasoning extends to the 4D‐Var objective function, when the estimation is taken over an assimilation window \([t_0, t_K]\), yielding
\begin{equation}
\begin{aligned}
J(x_{t_0})
= \frac{1}{2} \sum_{i=1}^K 
\bigl\|y_{t_i} - \mathcal{H}_{t_i}\bigl(\mathcal{M}_{0\to i}(x_{t_0})\bigr)\bigr\|^2_{\textbf{P}_{t_i}^{-1}}
+\frac{1}{2}\,\bigl\|x_{t_0}-x_{t_0}^b\bigr\|^2_{\textbf{B}_{t_0}^{-1}}.
\end{aligned}
\end{equation}
While assuming  
\(
R(x_t) \;=\; \frac{1}{2}\bigl\|x_t - x_t^b\bigr\|^2_{\textbf{B}_t^{-1}}
\)
yields a convenient quadratic objective, it is only a rough approximation of the true background error statistics.  In practice, strongly nonlinear dynamics and model imperfections as well as complex observational errors can induce non‐Gaussian errors. Accounting for such non-linear behaviors remains an active area of research in non‐Gaussian DA. In this paper, we embed rich, data‐driven priors via generative flow‐matching models learned from historical background-(re)analysis (as an approximation to the true state) pairs. We introduce briefly the OT Bayesian Flow Matching \cite{chemseddine2024conditional} in the following section.

\subsection{OT Bayesian Flow Matching}
Let $(\Omega,\mathcal F,\mathbb P)$ be a probability space and let 
$X^b,X^a\colon\Omega\to\mathcal X\subset\mathbb R^n$ 
be the background and analysis random fields, with realizations 
$x^b=X^b(\omega)$ and $x^a=X^a(\omega)$. Their joint distribution 
$P_{b,a}=(X^b,X^a)_\#\mathbb P\in\mathcal P_p(\mathcal X\times\mathcal X)$, 
where $\mathcal P_p(\mathcal X\times\mathcal X)$ denotes the set of probability measures on 
$\mathcal X\times\mathcal X$ with finite $p$th moments, has marginal 
$P_b=(X^b)_\#\mathbb P$. For a fixed background state $x^b\in\mathcal X$, the conditional probability of $X^a$ is 
$P_{a\mid b=x^b}=\mathrm{Law}(X^a\mid X^b=x^b)$, 
often written $p(X^a\mid X^b=x^b)$. To draw samples $x^a\sim P_{a\mid b=x^b}$, we construct a flow on the product space $\mathcal X\times\mathcal X$. Introducing an auxiliary variable 
$Z\colon\Omega\to\mathcal X$, independent of $X^b$ with law $P_Z$, we use the product measure 
$P_{b,Z}=P_b\otimes P_Z$ 
as the reference coupling for our flow-based sampler.

We denote by
\[
\Gamma\bigl(P_{b,a},\,P_{b,Z}\bigr)
\]
the set of all couplings—i.e.\ probability measures \(\alpha\) on \(\mathcal X^4\) whose marginal on \((x^b,x^a)\) is \(P_{b,a}\) and whose marginal on \((x^b,z)\) is \(P_{b,Z}\). To enforce that each plan preserves the background coordinate, we restrict to
\[
\Gamma^4_b
=\bigl\{\,
\alpha\in\Gamma(P_{b,a},P_{b,Z})
:\;
(\pi_{1,3})_\#\alpha=\Delta_\#P_b
\bigr\},
\]
where \(\pi_{1,3}\colon\mathcal X^4\to\mathcal X^2\) projects onto the first and third components, so \(\pi_{1,3}(x^b,x^a,(x^b)',z)=(x^b,(x^b)')\); for any measurable \(f\), \(f_\#\mu\) denotes the pushforward of \(\mu\); and \(\Delta\colon\mathcal X\to\mathcal X^2\) is the diagonal map \(\Delta(x^b)=(x^b,x^b)\), so \(\Delta_\#P_b\) is the law of \((X^b,X^b)\).
For each fixed background state $x^b\in\mathcal X$, the exact conditional Wasserstein-$p$ distance between $P_{b,a}$ and $P_{b,Z}$ is defined by
\begin{equation}
\begin{aligned}
W_{p,x^b}\bigl(P_{b,a},P_{b,Z}\bigr)
\;=\;
\inf_{\alpha\in\Gamma^4_b}
\Biggl(
\int_{\mathcal X^4}
\bigl\|\,(x^b,x^a)-\bigl((x^b)',z\bigr)\bigr\|^p
\;\mathrm{d}\alpha
\Biggr)^{\!1/p},
\end{aligned}
\label{eq:ExactCWD}
\end{equation}
where $\alpha = \alpha(x^b, x^a, (x^b)^\prime, z)$ and $\Gamma^4_b$ is as in the previous definition. In practice, the hard constraint $(x^b)'=x^b$ is relaxed by the penalized cost
\begin{equation}
\begin{aligned}
d_{p,\beta}\bigl((x^b,x^a),((x^b)',z)\bigr)
=
\|x^a - z\|^p
\;+\;
\beta\,\|x^b - (x^b)'\|^p,
\quad
\beta\gg1,
\end{aligned}
\end{equation}
leading to the approximate distance
\begin{equation}
\begin{aligned}
W_{p,\beta}
\;=\;
\inf_{\alpha\in\Gamma(P_{b,a},\,P_{b,Z})}
\Biggl(
\int_{\mathcal X^4}
d_{p,\beta}
\;\mathrm{d}\alpha
\Biggr)^{\!1/p},
\label{eq:ApproxCWD}
\end{aligned}
\end{equation}
which satisfies 
\[
\lim_{\beta\to\infty}W_{p,\beta}
\;=\;
W_{p,x^b}\bigl(P_{b,a},P_{b,Z}\bigr).
\]
Let \(\alpha^*\) denote the minimizer of the penalized problem in \eqref{eq:ApproxCWD}, and let
\[
T\colon \mathcal X^2 \to \mathcal X^2,
\qquad
T(x^b,z) = \bigl(x^b,\,T^a(x^b,z)\bigr)
\]
be the associated Monge map such that
\(
\alpha^* \;=\; (\mathrm{Id},T)_\#\bigl(P_b\otimes P_Z\bigr).
\)
Under this coupling \(\alpha^*\), one has almost surely\(\;T^a(x^b,z)=x^a\), where \(x^a=X^a(\omega)\). Using \(T^a(x^b,z)\) emphasizes that the analysis sample is obtained by transporting the auxiliary sample \(z\) under background \(x^b\), and to distinguish the Monge map from the abstract random field \(X^a\).
Along each fiber \(\{x^b\}\times\mathcal X\), set
\begin{equation}
X_\tau = (1-\tau)\,Z + \tau\,X^a,
\quad
\tau\in[0,1],
\label{eq:interpolant}
\end{equation}
so that \((x^b,x_t)\) traces the OT geodesic under \(\alpha^*\) with velocity 
\(
\dot x_\tau = \bigl(0,\;x^a - z\bigr).
\)
We then train the neural velocity field \(v_t^\theta\colon\mathcal X\times\mathcal X\to\mathbb R^{2n}\) by minimizing
\[
\mathbb{E}_{\substack{(x^b,z,x^b,x^a)\sim\alpha^*\\\tau\sim \mathcal U[0,1]}}\left[
\left\lVert
v_t^\theta\bigl(x^b,x_t\bigr)
-
\begin{pmatrix}0\\[0.3em]x^a - z\end{pmatrix}
\right\rVert^2\right],
\]
which is the original form of the loss averaged directly over the optimal plan \(\alpha^*\). At inference time, we solve
\[
\frac{\mathrm{d}\phi_\tau}{\mathrm{d} \tau}
= v_\tau^\theta(\phi_\tau),
\qquad
\phi_0=(x^b,z),
\]
so that 
\[
\phi_1=(x^b,x^a),
\quad
x^a\sim P_{a\mid b=x^b}.
\]
Thus we indeed recover independent samples from the conditional probability \(P_{a\mid b=x^b}\). Since the conditional‐Wasserstein plan \(\alpha^\ast\) enforces no transport in \(x^b\), ${x}^b$ is fixed and handled by the velocity networks conditional embedding, we can therefore omit the constant term and write the flow matching loss function as 
\begin{equation}
\mathbb{E}_{\substack{(x^b,z,x^b,x^a)\sim\alpha^\ast\\\tau\sim U[0,1]}}\left[
\Bigl\|
v_\tau^\theta\bigl(x^b,\,x_\tau\bigr)
-
\bigl(x^a - z\bigr)
\Bigr\|^2\right].
\label{eq:OTFMLoss}
\end{equation}
With the trained OT flow‐matching model, one can perform a purely generative correction of the background field \(x^b\) by drawing independent samples 
\(
x^a \sim P_{a\mid b=x^b}
\)
based on historical data. However, this model does not incorporate new observations into the state estimate and thus cannot perform true data assimilation on its own. As outlined in our motivation, we will instead use the flow‐matching model as a powerful prior within a plug‐and‐play (PnP) framework for solving the inverse problem \eqref{eq:GeneralInverseProblem}. In the following sections, we introduce \textsc{PnP‐DA}, where we adapt this PnP design to blend variational data assimilation with OT Bayesian flow matching, thereby combining observational corrections with an optimal‐transport‐based conditional generative prior. 
\subsection{Plug-and-Play Data Assimilation (\textsc{PnP‐DA})}
Proximal splitting methods are a set of iterative algorithms to solve inverse problems and have theoretical guarantee to converge to the minimizer of Equation \eqref{eq:GeneralInverseProblem} \cite{combettes2011proximal}, and can be used in DA for optimization as well \cite{ghosh2024robust,rao2017robust}. The proximal step requires an explicit $R(x)$, while Plug-and-Play (PnP) methods, first proposed by \cite{venkatakrishnan2013plug}, replace the proximal step with a denoising operator, and many of the denoising operators are neural networks to leverage rich data-driven priors. A brief review and convergence analysis can be found in \cite{ryu2019plug}. Recently, PnP methods have been successfully combined with modern flow‐matching and diffusion‐based generative models. These models offer expressive capacity to capture highly complex data distributions, an interactive sampling nature, and intrinsic connections to continuous‐time optimization, thus serving as effective denoising operators and enabling deeper integration into the PnP framework \cite{mardani2023variational,martin2024pnp,zhu2023denoising}. Among these generative priors, we build our data‐assimilation pipeline on PnP-Flow rather than on diffusion models. This choice is motivated by several theoretical advantages—most notably the straight-line flow used in flow matching \cite{martin2024pnp}, as opposed to the curved sampling trajectories in diffusion processes. We will illustrate these benefits in the following introduction of our \textsc{PnP‐DA} algorithm.

Given samples (in practice, mini-batches of samples) from the product distribution $P_Z\otimes P_{b,a}$, we can obtain the optimal conditional transport plan $\alpha^\ast$ and the corresponding Monge map $T$. Following the same argument in \cite{liu2022flow}, for the optimal coupling $\alpha^\ast$ and a background state $x^b$, given the straight-line flow setting defined in \eqref{eq:interpolant}, it is straightforward to see that the exact minimum of the OT Bayesian flow-matching target \eqref{eq:OTFMLoss} is achieved if

\begin{equation}
v^\ast_\tau(x^b, x)=\mathbb{E}\left[
X^a - Z \mid X^b = x^b, X_\tau=x 
\right],
\label{velocity}
\end{equation}
where the expectation is taken over the line directions $X^a-Z$ on the fiber \(\{x^b\}\times\mathcal X\). We define the Bayesian extension of the standard OT flow denoising operator used in \cite{martin2024pnp} as following: for any $x\in\mathbb{R}^d$ and $\tau\in[0,1]$,

\begin{equation}
    D_\tau(x^b, x)=x+(1-\tau)v_\tau^\ast(x^b,x).
\label{denoising operator}
\end{equation}
The theoretical insight of the denoising operator can be seen from the following: with a properly trained neural velocity field where $v_\tau^\theta=v_\tau^\ast$, from \eqref{eq:interpolant}, \eqref{velocity} and \eqref{denoising operator}, we have 
\begin{equation}
\begin{aligned}
D&_\tau(x^b, x) \\ 
=x&+(1-\tau)v_\tau^\ast(x^b,x)\\
=\mathbb{E}&\left[X_\tau\mid X^b=x^b, X_\tau=x\right]  + (1-\tau)\mathbb{E}\left[
X^a - Z \mid X^b = x^b, X_\tau=x 
\right]\\
=\mathbb{E}&\left[X^a\mid X^b = x^b, X_\tau=x  \right],
\label{denoising operatorInterp}
\end{aligned}
\end{equation}
where the expectations are taken over the line directions $X^a-Z$ on the fiber \(\{x^b\}\times\mathcal X\). Therefore, our OT Bayesian flow‐matching denoiser provides the optimal estimate of \(X^a\) conditioned on the current flow state \(X_\tau = x\) and the background field \(X^b = x^b\). We integrate this denoising step with the other two operations of the PnP-Flow algorithm to perform data assimilation (see Algorithm \ref{alg:pnp_cmf_da}). 

\begin{algorithm}
\caption{PnP-DA Analysis step (3D)}  
\label{alg:pnp_cmf_da}
\begin{algorithmic}[1]
  \Require Observations $y_t$, background state $x_t^b$, observation operator $\mathcal{H}_t$,  
           observation error covariance $\textbf{P}_t$, denoiser $D_\tau$,  
           pseudo-time sequence $(\tau_n)_n$, learning-rates $(\gamma_n)_n$
  \State Initialize $x^{(0)} \gets x_t^b$
  \For{$n=0,1,\dots$}
    \State $w^{(n)} \gets x^{(n)} - \gamma_n \nabla\!\Bigl(\tfrac12\|y_t-\mathcal H_t(x^{(n)})\|^2_{\textbf{P}_t^{-1}}\Bigr)$
    \State $\tilde w^{(n)} \gets (1-\tau_n)z + \tau_n\,w^{(n)},\;z\sim P_Z$
    \State $x^{(n+1)} \gets D_{\tau_n}(x_t^b, \tilde w^{(n)})$
  \EndFor\\
  \Return Analysis state $x_t^a$
\end{algorithmic}
\end{algorithm}

\textsc{PnP‐DA} first performs a gradient descent on the observation misfit to nudge the state toward new measurements. The misfit is quantified via the Mahalanobis distance under an unknown distribution characterized by a positive semi-definite covariance matrix \(\textbf{P}_t\), which ensures each observation influences the analysis in proportion to its uncertainty and balances information content against error characteristics. That gradient update is immediately blended with a stochastic draw from a reference distribution and the current state, projecting the estimate onto a learned flow path. A conditional denoising step then enforces regularization along that trajectory. This tightly coupled design yields an informative prior—shaped by historical analyses conditioned on the current background—while the gradient step simultaneously ensures fidelity to incoming observations. Also, \textsc{PnP‐DA} never backpropagates gradient through the pre-trained network—instead, it confines all differentiation to the observation misfit and applies the denoiser only via a single forward pass per iteration. Therefore, \textsc{PnP‐DA} avoids unstable gradient propagation through complex neural networks and reduces computational overhead, thereby improving stability and accelerating the analysis step. 

\section{Numerical Experiments and Results}
Operational forecasting systems invariably contain errors—from imperfect parameters and unresolved physical processes to stochastic perturbations and uncertain initial conditions—while observations of the true system remain sparse and noisy. To emulate these dual challenges and demonstrate \textsc{PnP‑DA}’s ability, we employ three idealized testbeds that serve as standard benchmarks in the data‑assimilation literature (e.g.,\cite{tamang2021ensemble,law2016filter,evensen2024iterative, protas2004computational}).

\paragraph{Lorenz 63 system} The Lorenz 63 model \cite{DeterministicNonperiodicFlow} arises from a three-mode truncation of the Fourier expansion of the Rayleigh-Bénard convection equations, yielding a system described as follows:
\begin{equation}
\begin{aligned}
    \frac{dX}{dt}&=-\sigma(X-Y),\\
    \frac{dY}{dt}&=\rho X - Y- XZ,\\
    \frac{dZ}{dt}&=XY-\beta Z.
\end{aligned}
\label{eq:l63}
\end{equation}
For the “true” system, we use $\sigma=10$, $\rho=28$ and $\beta=8/3$, and the system exhibit chaotic behavior, making it a classic testbed for studying predictability in simplified atmospheric models. To introduce systematic error into an "operational" model, we perturb the parameters to $\sigma^\prime=10.5$, $\rho^\prime=27$, and $\beta^\prime=10/3$, and to model random error we add noise  $\omega\sim\mathcal{N}(0,0.02\mathbf{I})$ at each time step, following numerous DA studies like \cite{tamang2021ensemble, amezcua2014ensemble}.

\paragraph{Lorenz 96 system} The Lorenz 96 system \cite{lorenz1996predictability} is a two-scale dynamical system that captures the essential nonlinear advection and multiscale coupling characteristic of extra-tropical atmospheric flow. In our experiments, we consider a scenario that mirrors a common challenge in Earth system modeling: the presence of unresolved physics that leads to deviation of the model from the truth. We use a two-scale Lorenz 96 model as the ``true'' system, which is governed by the following equations:

  \begin{equation}
  \begin{aligned}
      \frac{dX_k}{dt} &= -X_{k-1}(X_{k-2} - X_{k+1}) - X_k + F + \frac{hc}{b}\sum_{j=1}^J Y_{j,k},\\
      \frac{dY_{j,k}}{dt} &= -cb\,Y_{j+1,k}(Y_{j+2,k} - Y_{j-1,k}) - c\,Y_{j,k} + \frac{hc}{b}\,X_k.
  \end{aligned}
  \end{equation}
where $X_k$ represents the slow variables for $k = 1,\ldots,K$ and $Y_{j,k}$ represents the fast variables for $j = 1,\ldots,J$. We set $K = 8$, $J = 32$, $F = 18$, $h = 1$, $b = 10$, and $c = 10$, following established parameter values to generate chaotic behavior \cite{balwada2024learning}. To simulate this scenario where certain physical processes are unrepresented, we use a single-scale Lorenz 96 model as our operational system:

\begin{equation}
    \frac{dX_k}{dt} = -X_{k-1}(X_{k-2} - X_{k+1}) - X_k + F',
\end{equation}
where $F'$ is a modified forcing parameter that attempts to account for the missing fast-scale dynamics. 

\paragraph{Kuramoto-Sivashinsky (KS) equation} The KS equation is a fourth-order one-dimensional nonlinear partial differential equation that models instabilities in laminar flame fronts and dissipative systems with competing dispersion and dissipation \cite{kuramoto1978diffusion,SIVASHINSKY1988459}.  Although not directly related to Earth system modeling, the equation is well known for its spatiotemporal chaos that shares features with turbulence and real geophysical fluid dynamics. Therefore, the KS equation serves as a canonical testbed for high-dimensional data assimilation \cite{evensen2024iterative}. The KS equation reads:

\begin{equation}
\frac{\partial u}{\partial t} = -u \frac{\partial u}{\partial x} - \frac{\partial^2 u}{\partial x^2} - \nu\frac{\partial^4 u}{\partial x^4},
\label{eq:ks}
\end{equation}

where $u(x,t)$ is a scalar field defined on $x\in[0,L]$, $t\in[0,\infty]$, and $\nu$ is a scalar coefficient set to be $0.5$ in our experiments. We impose the Dirichlet boundary condition $u(0,t)=u(L,t)=0$. We also introduce noise in the operational model by adding noise $\omega\sim\mathcal{N}(0,0.01\mathbf{I})$.

\subsection{Data and Training}
We generate a nature run by integrating the true system with a fourth-order Runge–Kutta scheme (RK4), then discard its initial transient to ensure the attractor is reached. 
\begin{itemize}
  \item \textbf{Lorenz 63:} time step $\Delta t = 0.01$; observations every $\Delta t_{\mathrm{obs}} = 0.4$ with additive zero-mean Gaussian noise. 
  \item \textbf{Two‑scale Lorenz 96:} time step $\Delta t = 0.005$; we observe only 50\% of the slow variables $X_k$ at interval $\Delta t_{\mathrm{obs}} = 0.2$ (about 2.5 days), each observation perturbed by
    $\mathcal{N}(0,\,\sigma_{\mathrm{obs}}^2)$ with $\sigma_{\mathrm{obs}} = 0.5$.
  \item \textbf{KS Equation:} time step $\Delta t=0.25$, domain length $L=50$ discretized into $128$ grids; we observe only $16$ equally spaced grid points at interval $\Delta t_{\mathrm{obs}}=2.5$, each observation perturbed by $\mathcal{N}(0,\,\sigma_{\mathrm{obs}}^2)$ with $\sigma_{\mathrm{obs}} = 0.1$.
\end{itemize}

To create training data for our OT Bayesian flow matching model, we first generate a set of background-reanalysis pairs. The background states are produced by running the operational systems as described in the previous section from previous analysis states. The corresponding analysis states are obtained by running cyclic data assimilation to assimilate sparse and noise observations of the true system into the operational models. For the Lorenz-63 system we use the Ensemble Riemannian data assimilation (EnRDA) with 10 ensemble members, for the Lorenz 96 system we use the Ensemble Kalman Filter (EnKF) with 20 ensemble members, and for the KS equation we use EnKF with 40 ensemble members. This process yields paired data $(x^b, x^a)$ consisting of background forecasts and their corresponding reanalysis states. These pairs capture the systematic relationship between the imperfect model forecasts and the best estimate of the true state given the available observations. Our OT Bayesian flow matching model learns from these pairs, effectively learning to correct the systematic errors introduced by model simplification given certain observations and their statistics. Further details of numerical simulations and dataset can be found in the supplementary materials.

\paragraph{Implementation and Training} Our conditional flow network is a multilayer perceptron that predicts the OT Bayesian flow‑matching velocity $v_\tau^\theta(x^b, x)$ takes the current flow matching state $x$, the background forecast $x^b$ and pseudo-time $\tau$ as a form of a Fourier embedding. The core of the model consists of four hidden layers, and residual connections wherever a block’s input and output widths match. A final linear layer projects back to the state dimension, yielding the velocity vector. Full details like hyperparameter settings are provided in the supplementary materials.

\subsection{Results}
\label{subsec:results}

\paragraph{Baselines} We compare our \textsc{PnP‐DA} method against 3D-Var using its optimal analytical solution (i.e., Eq \eqref{eq:3dvar-sol}). The background covariance matrix for 3D-Var is estimated using the Gaspari-Cohn function \cite{gaspari1999construction}. We ran 50 independent cyclic DA experiments for each method on each testbed. For \textsc{PnP-DA} we use 100 iterations and $\gamma_\tau=(1-\tau)^\alpha$ with $\alpha=0.5$, based on the setting used in \cite{martin2024pnp}. We have conducted ablation study on both of the number of iterations and the choice of $\alpha$,  demonstrating that as few as 10 iterations suffice to surpass the 3D‑Var baseline in mean RMSE. Full results and discussion are provided in the supplementary material.

\begin{figure}[h]
\centering
\centerline{\includegraphics[width=\columnwidth]{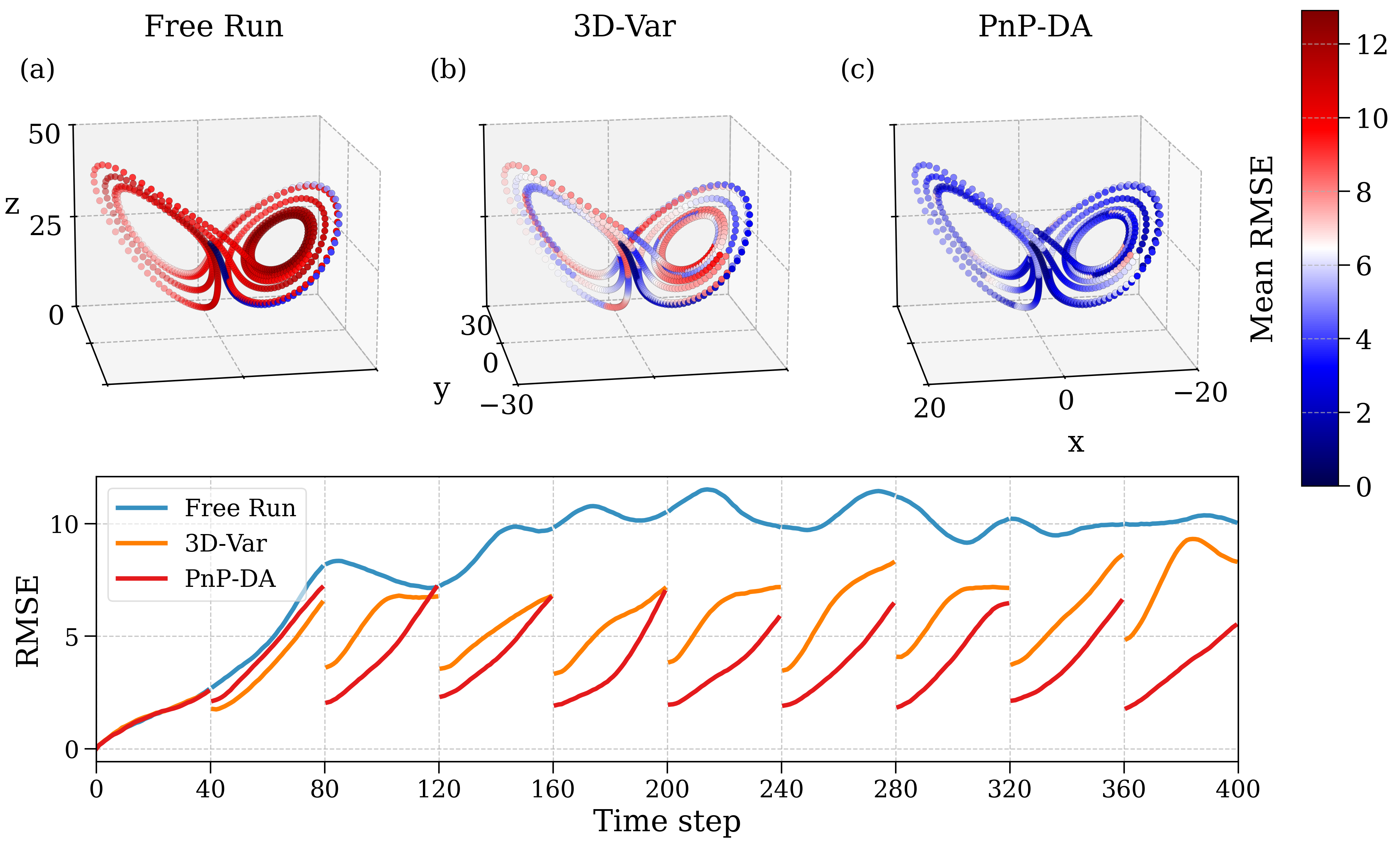}}
\caption{Temporal evolution of RMSE of the Lorenz 63 system in phase space (upper panel) and time (lower panel).}
\label{fig:lorenz633d}
\end{figure}

\begin{table}[h]
  \centering
  
  \begin{tabular}{lccc}
    \toprule
    Method   & $x$        & $y$        & $z$        \\
    \midrule
    Free Run & 8.31$\pm1.57$      & 9.35$\pm1.64$      & 8.75 $\pm1.80$     \\
    3D‑Var   & 4.61$\pm1.85$      & 6.74$\pm2.27$     & 5.13$\pm1.36$     \\
    PnP‑DA   & \textbf{2.54$\pm0.79$}      & \textbf{3.94$\pm1.18$}      & \textbf{3.58$\pm1.01$}      \\
    \bottomrule
  \end{tabular}
  \caption{RMSE of Lorenz 63 experiments. }
  \label{tbl:rmse_xyz}
\end{table}

\paragraph{Lorenz 63}  For Lorenz 63, only perturbed $x$ and $z$ are observed every 40 time steps. The time evolution of the step-wise root mean square error (RMSE) averaged over 50 experiments is colored over the phase space, and also plotted in time series in Figure \ref{fig:lorenz633d}. The mean RMSE calculated over the whole trajectory for each dimension, along with standard deviation, are presented in Table \ref{tbl:rmse_xyz}.  We can see the improvement of \textsc{PnP-DA} compared to 3d-Var is obvious. While the training data is generated with observing all variables at each observation step, as required by EnRDA, our approach generalizes well when the observation become incomplete. Further visualizations of the time series of trajectories can be found in supplementary material.
\begin{figure}[h]
\centering
\centerline{\includegraphics[width=\columnwidth]{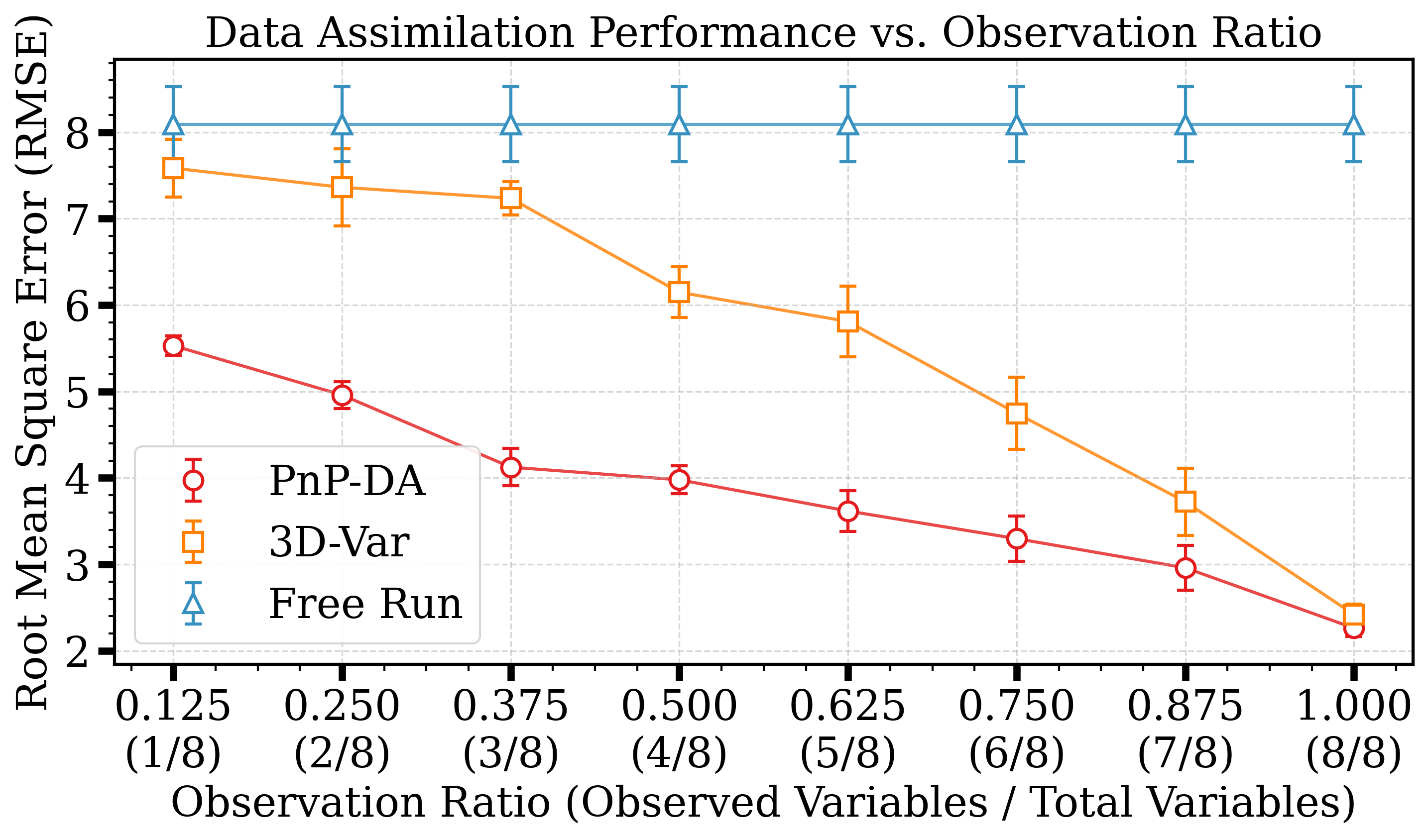}}
\caption{RMSE with different number of observations.}
\label{fig:rmse_comparison}
\end{figure}

\begin{figure}[h]
\centering
\centerline{\includegraphics[width=\columnwidth]{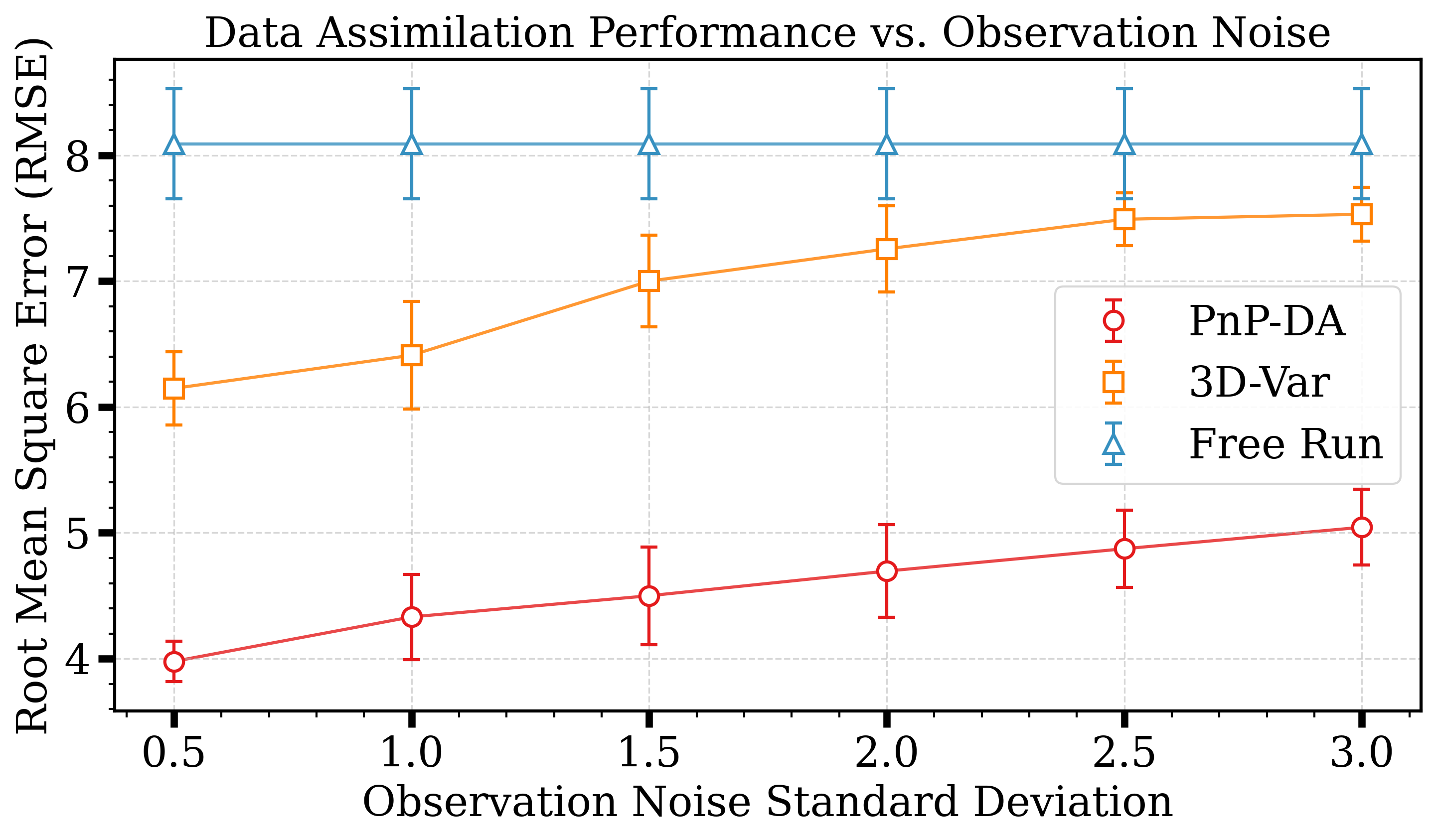}}
\caption{RMSE with different noise levels.}
\label{fig:noise_comparison}
\end{figure}

\paragraph{Lorenz 96} Given a more challenging task that there are missing processes in the operational model, we further stress test our approach by making observations more sparse and noisy. Figure \ref{fig:rmse_comparison} shows the RMSE over the whole trajectory at varying observation densities. Our \textsc{PnP‐DA} method consistently outperforms 3D-Var approach, particularly in observation-sparse regimes where only 10-30\% of the variables are observed. The performance gap persists as observations become more sparse, highlighting the value of the conditional prior in regimes where the data fidelity term provides less constraint. With full observations, all methods perform similarly, as the solution becomes increasingly determined by the observations rather than the prior. Figure \ref{fig:noise_comparison} shows the mean RMSE as the observation noise standard deviation $\sigma_{\mathrm{obs}}$ increases from 0.5 to 3.0. At $\sigma_{\mathrm{obs}}=0.5$, \textsc{PnP‐DA} achieves an RMSE near 4.0, compared to $\approx 6.1$ for 3D-Var and $\approx 8.1$ for the free run. As noise grows, 3D-Var’s error climbs sharply to about 7.5 at $\sigma_{\mathrm{obs}}=3.0$, whereas \textsc{PnP‐DA} degrades more gradually to roughly 5.0, preserving a 30–35\% advantage. The sustained gap between \textsc{PnP‐DA} and 3D-Var under rising $\sigma_{\mathrm{obs}}$ highlights the robustness of our \textsc{PnP‐DA} approach against increasing observation noise levels.  

\begin{figure}[h]
\centering
\centerline{\includegraphics[width=\columnwidth]{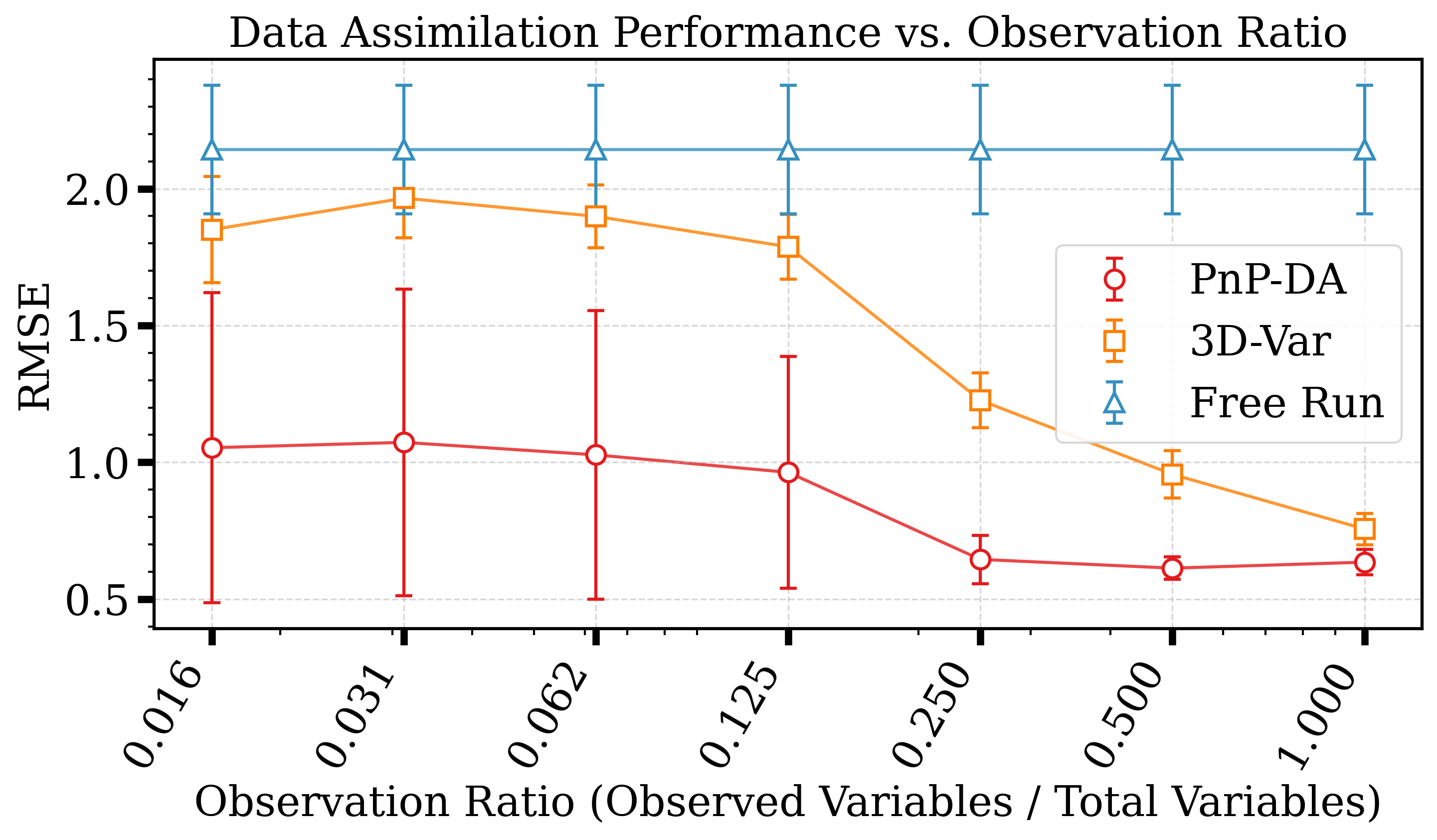}}
\caption{RMSE vs Observation Ratio for KS.}
\label{fig:ks_obs_ratio}
\end{figure}

\begin{figure}[h]
\centering
\centerline{\includegraphics[width=\columnwidth]{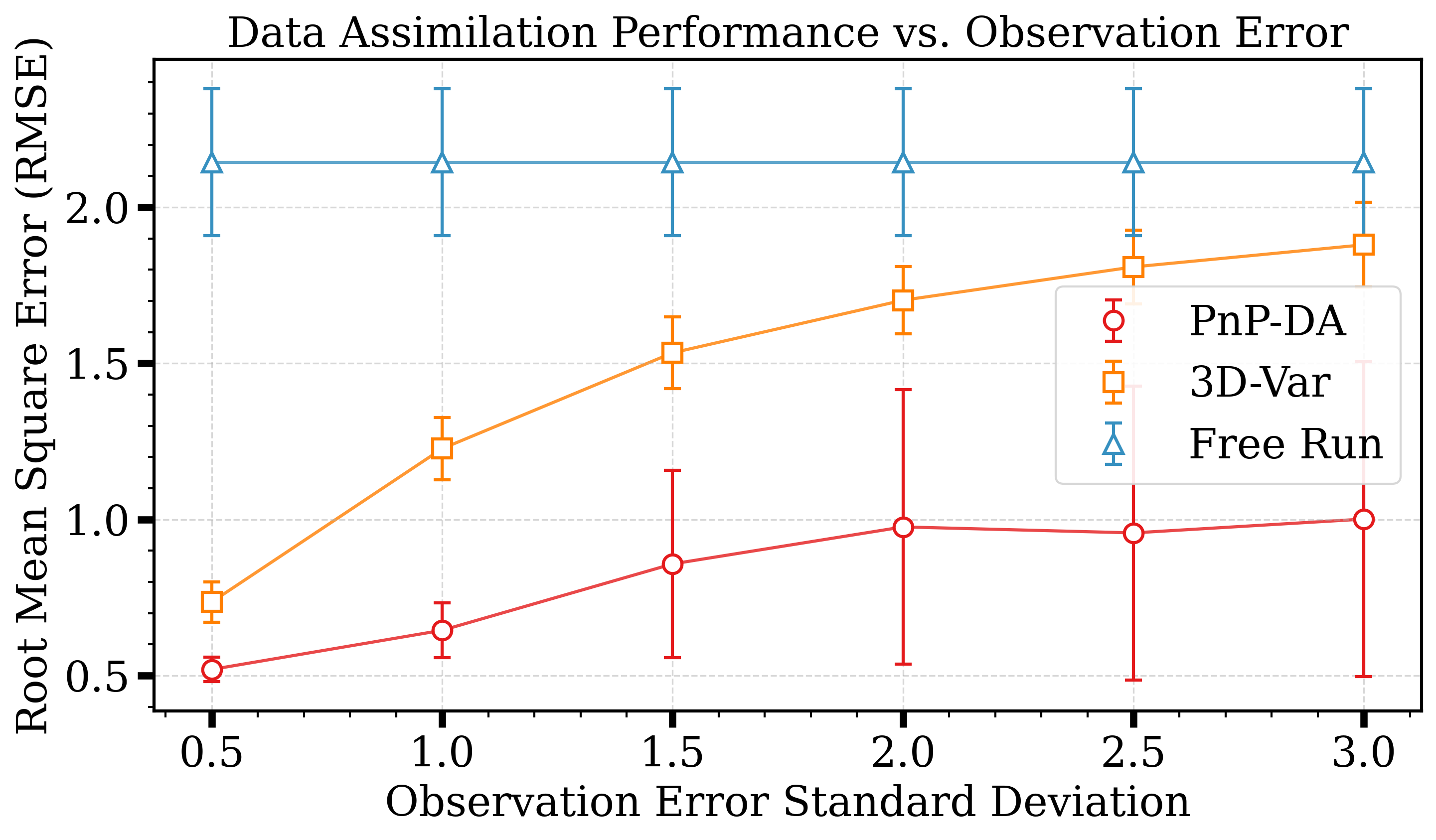}}
\caption{RMSE vs Observation Error Level for KS.}
\label{fig:ks_error_std}
\end{figure}

\paragraph{KS Equation} Using this infinite-dimensional dynamical system we can test more extreme cases when the observation ratio 10 times lower that that in Lorenz 96 cases.  Figure~\ref{fig:ks_obs_ratio} (\(\sigma_{\text{obs}}=1\)) and Figure~\ref{fig:ks_error_std} (32 observations) show that \textsc{PnP‑DA} continues to outperform 3D-Var across observation densities and noise levels. The mean performance gap remains substantial even in this more challenging regime, demonstrating the method’s scalability and robustness.
Compared to Lorenz 96, we observe a higher variance in the RMSE across runs. This may be attributed to the KS system’s increased sensitivity to perturbations, numerical stiffness in the solver, and lack of dedicated hyperparameter tuning. Despite this, the mean performance of \textsc{PnP‑DA} remains consistently superior. These results further support the advantage of incorporating learned priors into variational data assimilation, particularly in ensemble settings where we perform an ensemble of DA and use the mean as the analysis. With randomness in the sampling process in \textsc{PnP-DA} we can easily generate analysis ensembles without performing ensemble forecasting, contrary to classical ensemble variational DA.

\section{Discussion and Conclusion}

Our experiments on the testbed ranging from idealized earth system model to infinite-dimensional dynamical system demonstrate that \textsc{PnP‐DA}—by alternating a lightweight gradient‐descent update on the observation misfit with a single forward pass through a pretrained OT Bayesian flow‐matching denoiser conditioned via a conditional Wasserstein coupling—effectively outperforms its classical variational counterpart. In observation-sparse regimes, \textsc{PnP‐DA} significantly reduces forecast error, and under increasing noise, it degrades more gracefully, maintaining a substantial accuracy margin. These improvements arise from the denoiser’s ability to incorporate error‐correction patterns learned from historical background–analysis pairs without requiring expensive backpropagation during assimilation, thus ensuring numerical stability and computational efficiency.

Moreover, the \textsc{PnP‐DA} pipeline is highly flexible: the same pretrained flow‐matching prior is used unchanged for both 3D and 4D assimilation windows (see supplementary materials), and the conditional coupling enforces minimal transport of the background forecast \(x^b\) while optimally generating samples from the conditional analysis distribution. By restricting gradient computations to the observation misfit and using the pretrained network in only a single forward pass per iteration, \textsc{PnP‐DA} sidesteps unstable gradient flows in complex models, slashes computational overhead, and accelerates the analysis step. This integration of Bayesian OT flows with interpretability and generative modeling offers a powerful new paradigm for data assimilation. 
\paragraph{Limitation and Future Research} Our experiments so far have been confined to the Lorenz systems and KS equation. Demonstrating that \textsc{PnP‐DA} scales to high-dimensional, operational Earth system models will require further testing, and we are pursuing a formal convergence analysis. Looking ahead, we plan to extend our work to 4D assimilation settings, benchmark \textsc{PnP‐DA} against established methods in realistic NWP applications—both atmospheric and oceanic—and incorporate model error reduction \cite{qu2023can, li2024machine, lex} with historical data/high-resolution simulations and real world observations. Finally, recent advances in physics-constrained generative modeling offer compelling strategies for embedding conservation laws directly into the sampling process \cite{blanke2025strictly, cheng2025gradient}; given the critical role of mass, energy, and momentum conservation in geophysical flows \cite{ruckstuhl2021training}, imposing physical constraints within the \textsc{PnP‐DA} workflow represents a promising avenue for future research.

\paragraph{Acknowledgement:}
We acknowledge funding from NSF through the Learning the Earth with Artificial Intelligence and Physics (LEAP) Science and Technology Center (STC) (Award \#2019625) and USMILE European Research Council synergy grant. We would like to acknowledge high-performance computing support from the Derecho system (doi:10.5065/qx9a-pg09) provided by the NSF National Center for Atmospheric Research (NCAR), sponsored by the National Science Foundation.

\bibliographystyle{abbrv}
\bibliography{references}

\appendix
\section{Experimental Setup}
\paragraph{Lorenz‑63 Data Generation}

\begin{enumerate}
  \item \textbf{Nature Run:}
  \begin{itemize}
    \item Integrate the true Lorenz‑63 system
    \[
      \frac{dx}{dt} = -\sigma\,(x - y),\quad
      \frac{dy}{dt} = \rho\,x - y - x\,z,\quad
      \frac{dz}{dt} = x\,y - \beta\,z,
    \]
    with parameters \(\sigma=10,\;\rho=28,\;\beta=8/3\).
    \item Use 4th‑order Runge–Kutta (RK4) with step size \(\Delta t=0.01\).
    \item Spin‑up for 5000 steps, discard first 1000
    \item Record trajectory \(x_t\in\mathbb R^3\) for \(t=0,\ldots,T\) with \(T=100{,}000\).
  \end{itemize}

  \item \textbf{Synthetic Observations:}
  \begin{itemize}
    \item Observation times \(t_k = k \times 40\,\Delta t\) (every 40 steps, \(\Delta t_{\mathrm{obs}}=0.4\)).
    \item Generate noisy observations
    \[
      y_k = x_{t_k} + \varepsilon_k,\quad
      \varepsilon_k \sim \mathcal{N}(0,\Sigma_{\mathrm{obs}}),
    \]
    with \(\Sigma_{\mathrm{obs}} = \sigma_{\mathrm{obs}}^2\,C\), \(\sigma_{\mathrm{obs}}=\sqrt{2}\), and
    \[
      C = \begin{pmatrix}
            1 & 0.5 & 0.25\\
            0.5 & 1 & 0.5\\
            0.25 & 0.5 & 1
          \end{pmatrix}.
    \]
  \end{itemize}

  \item \textbf{Ensemble Riemannian DA (EnRDA) Loop: \cite{tamang2021ensemble}}
  \begin{itemize}
    \item Ensemble size \(N=10\). Initialize
    \[
      x_0^{(i)} = x_0 + \delta^{(i)},\quad
      \delta^{(i)}\sim\mathcal{N}(0,\sigma_{\mathrm{init}}^2 I_3),\;
      \sigma_{\mathrm{init}}=\sqrt{2}.
    \]
    \item \emph{Forecast step}: each member is propagated under a perturbed \emph{operational} model with parameters
    \[
      (\sigma',\rho',\beta')=(10.5,\,27.0,\,10/3)
    \]
    via
    \[
      x_{t+1}^{b,(i)} = \mathrm{RK4}(x_t^{a,(i)};\Delta t)
      + \eta_t^{(i)},\quad
      \eta_t^{(i)}\sim\mathcal{N}(0,\,2\Delta t\,I_3).
    \]
    \item \emph{Analysis step} at \(t_k\):
    \begin{enumerate}
      \item Form perturbed observation ensemble \(y_k^{(i)} = y_k + \delta_k^{(i)}\), \(\delta_k^{(i)}\sim\mathcal{N}(0,\Sigma_{\mathrm{obs}})\).
      \item Compute background covariance \(B = \mathrm{Cov}(\{x_{t_k}^{b,(i)}\})\).
      \item Compute entropic OT plan with Sinkhorn iterations
      \[
        U = \mathrm{entrop\_OMT}\bigl(X,Y,\gamma=10,n_{\mathrm{iter}}=300\bigr).
      \]
      \item Mixture update for each member:
      \[
        x_{t_k}^{a,(i)} = \eta\,x_{t_k}^{b,(I_i)} + (1 - \eta)\,y_k^{(J_i)},\quad
        \eta = \frac{\operatorname{tr}(\Sigma_{\mathrm{obs}})}{\operatorname{tr}(\Sigma_{\mathrm{obs}})+\operatorname{tr}(B)},
      \]
      where \((I_i,J_i)\) are sampled according to weights \(U\).
    \end{enumerate}
  \end{itemize}
\end{enumerate}

\paragraph{Lorenz‑96 Data Generation}

\begin{enumerate}
  \item \textbf{Nature Run:}
  \begin{itemize}
    \item Integrate the two‑scale Lorenz‑96 system
      \[
        \frac{dX_k}{dt} = -X_{k-1}\,(X_{k-2}-X_{k+1}) - X_k + F + \frac{h\,c}{b}\sum_{j=1}^J Y_{j,k},
      \]
      \[
        \frac{dY_{j,k}}{dt} = -c\,b\,Y_{j+1,k}\,(Y_{j+2,k}-Y_{j-1,k}) - c\,Y_{j,k} + \frac{h\,c}{b}\,X_k,
      \]
      with \(K=8\), \(J=32\), \(F=18\), \(h=1\), \(b=c=10\).
    \item Use 4th‑order Runge–Kutta with \(\Delta t = 0.005\); spin‑up 10,000 steps, discard first 2,000.
  \end{itemize}

  \item \textbf{Synthetic Observations:}
  \begin{itemize}
    \item Observe 50\% of slow variables \(X_k\) every
      \(\Delta t_{\mathrm{obs}} = 0.2\).
    \item Corrupt with Gaussian noise
      \[
        y_k = X_{k}(t_k) + \varepsilon_k,\quad
        \varepsilon_k \sim \mathcal{N}(0,\sigma_{\mathrm{obs}}^2),\;
        \sigma_{\mathrm{obs}} = 0.5.
      \]
  \end{itemize}

  \item \textbf{Cyclic EnKF Assimilation (N=20):}
  \begin{itemize}
    \item \emph{Forecast:} Propagate each analysis member \(x^{a,(i)}_{t_{k-1}}\) under the single‑scale Lorenz‑96 model
      \[
        \frac{dX_k}{dt} = -X_{k-1}\,(X_{k-2}-X_{k+1}) - X_k + F',
      \]
      with \(F'=18\), over \(\Delta t_{\mathrm{obs}}\) to obtain background ensemble \(\{x^{b,(i)}_{t_k}\}\).
    \item \emph{Analysis:}
      \begin{itemize}
        \item Compute background mean \(\bar x^b\) and covariance
          \(\;B = \frac{1}{N-1}\sum_i (x^{b,(i)}-\bar x^b)(x^{b,(i)}-\bar x^b)^\top.\)
        \item Kalman gain:
          \[
            K = B\,H^\top\bigl(H\,B\,H^\top + R\bigr)^{-1},\quad
            R = \sigma_{\mathrm{obs}}^2 I.
          \]
        \item Update each member:
          \[
            x^{a,(i)}_{t_k} = x^{b,(i)}_{t_k} + K\bigl(y^{(i)}_{t_k} - H\,x^{b,(i)}_{t_k}\bigr).
          \]
      \end{itemize}
    \item Store \(\bar x^b_{t_k}\) and \(\bar x^a_{t_k} = \tfrac1N\sum_i x^{a,(i)}_{t_k}\) as training pairs \((x^b,x^a)\).
  \end{itemize}
\end{enumerate}

\paragraph{KS Equation Data Generation}
\begin{enumerate}
  \item \textbf{Nature Run:}
  \begin{itemize}
    \item Integrate the Kuramoto–Sivashinsky equation
      \[
        \frac{\partial u}{\partial t} = -u\,\frac{\partial u}{\partial x} - \frac{\partial^2 u}{\partial x^2}
          - \nu\,\frac{\partial^4 u}{\partial x^4},
      \]
      on \(x\in[0,L]\) with Dirichlet BCs \(u(0)=u(L)=0\), where \(\nu=0.5\), \(L=50\).
    \item Discretize into \(n_e=128\) interior points; use LSODA solver with \(\Delta t=0.25\).
    \item Spin‑up for \(t_{\rm ini}=50\) (ns=200 steps), then record true state \(u_{\rm true}(x,t)\) for \(t\in[0,100]\) (nt=400 steps).
  \end{itemize}

  \item \textbf{Synthetic Observations:}
  \begin{itemize}
    \item Observe \(m_e=16\) equally spaced grid points at times \(t_b=k\,(t_{\max}/n_f)\), where \(n_f=10\).
    \item Add independent noise
      \[
        u_{\rm obs} = u_{\rm true}(x_{\rm obs},t_b) + \varepsilon,\quad
        \varepsilon\sim\mathcal{N}(0,\sigma_{\rm obs}^2),\;\sigma_{\rm obs}=\sqrt{10^{-2}}.
      \]
  \end{itemize}

  \item \textbf{Cyclic EnKF Assimilation (N=40):}
  \begin{itemize}
    \item \emph{Forecast:} For each of \(N=40\) members, integrate the “bad” forecast model (same KS but adding process noise)
      \[
        u^{b,(i)}_{t+\Delta t} = \text{LSODA}\bigl(u^{a,(i)}_t\bigr) + \eta,\quad
        \eta\sim\mathcal{N}(0,\sigma_{\rm mod}^2I),\;\sigma_{\rm mod}=\sqrt{10^{-2}}.
      \]
    \item \emph{Analysis:} At each \(t_b\):
      \begin{itemize}
        \item Perturb observations: \(u^{o,(i)} = u_{\rm obs} + \delta^{(i)},\;\delta^{(i)}\sim\mathcal{N}(0,\sigma_{\rm obs}^2I)\).
        \item Compute background mean \(\bar u^b\) and covariance \(B\) over ensemble.
        \item Compute Kalman gain
          \[
            K = B\,H^\top\bigl(H\,B\,H^\top + R\bigr)^{-1},\quad
            R = \sigma_{\rm obs}^2 I_{m_e}.
          \]
        \item Update each member:
          \[
            u^{a,(i)} = u^{b,(i)} + K\bigl(u^{o,(i)} - H\,u^{b,(i)}\bigr).
          \]
      \end{itemize}
    \item Store \(\bar u^b\) and \(\bar u^a\) (ensemble means) as training pairs.
  \end{itemize}
\end{enumerate}

\subsubsection{Training Settings}

    \begin{itemize}
      \item Optimizer: AdamW, \(\mathrm{lr}=3\times10^{-4}\), weight decay \(10^{-5}\).
      \item Scheduler: ReduceLROnPlateau (factor 0.5, patience 10).
      \item Batch size: 32; epochs: 1000 but with early stopping on validation loss (patience 50).
    \end{itemize}

\subsubsection{Neural Network Architecture}
\begin{itemize}
  \item \textbf{Time Embedding:}
    \begin{itemize}
      \item Fourier‐based embedding of scalar \(t\in[0,1]\) into \(\mathbb R^{d_\tau}\) with
        \[
          \mathrm{angles} = 2\pi\,t\,G^\top,\quad G\in\mathbb R^{(d_\tau/2)\times1}, 
        \]
        where \(d_\tau=32\) and \(G\) is a fixed Gaussian random matrix scaled by 10.
      \item Output: \(\bigl[\sin(\mathrm{angles}),\;\cos(\mathrm{angles})\bigr]\in\mathbb R^{d_\tau}.\)
    \end{itemize}

  \item \textbf{Conditional Flow Network:}
    \begin{itemize}
      \item Inputs: current state \(x_t\in\mathbb R^d\), time‐embed \(\in\mathbb R^{d_\tau}\), background \(x^b\in\mathbb R^d\).
      \item Concatenate to \(z_0\in\mathbb R^{2d+d_\tau}\).
      \item MLP backbone:
        \begin{itemize}
          \item Layers: (Lorenz 63 [32, 64, 64, 32], Lorenz 96 [32, 64, 128, 64, 32], KS Equation [256, 256,512, 256, 256] )hidden units.
          \item Activation: SiLU; LayerNorm after each hidden layer.
          \item Residual blocks when input and output dims match.
        \end{itemize}
      \item Final linear layer projects to \(\mathbb R^d\) to predict vector field \(v_\theta(z_t,t,x^b)\).
    \end{itemize}
\end{itemize}

\paragraph{Bayesian OT Flow‑Matching Loss \cite{chemseddine2024conditional}}
Given a minibatch of \(N\) paired samples \(\{(x^b_i,x^a_i)\}_{i=1}^N\), we denote
\[
  z_0^{(i)}\sim\mathcal{N}(0,I_d),\quad
  t^{(i)}\sim\mathcal{U}[0,1],\quad
  x^b_i,x^a_i\in\mathbb{R}^d.
\]
If Bayesian OT is enabled (we use \(\beta=1000\)), we form augmented source and target vectors
\[
  s_i = \bigl(z_0^{(i)},\,\sqrt\beta\,x^b_i\bigr),\quad
  r_j = \bigl(x^a_j,\,\sqrt\beta\,x^b_j\bigr)\quad(\in\mathbb{R}^{2d}),
\]
compute the squared‐cost matrix 
\(\,C_{ij}=\|s_i - r_j\|^2\), and solve the Earth Mover’s Distance plan \(\pi\in\mathbb R^{N\times N}\) via  
\(\mathrm{emd}(\,\tfrac1N\mathbf1,\tfrac1N\mathbf1,C)\).  
We then define the target velocity for each \(i\) by
\[
  v^*_t{}^{(i)} = x^a_{j(i)} - z_0^{(i)},\quad
  j(i)=\arg\max_j \pi_{ij}\,.
\]
Otherwise (\(\beta=0\)), we simply set 
\(\;v^*_t{}^{(i)} = x^a_i - z_0^{(i)}.\)

We interpolate 
\[
  z_t^{(i)} = z_0^{(i)} + t^{(i)}\,v^*_t{}^{(i)},
\]
and predict 
\(\;v_\theta^{(i)} = f_\theta\bigl(z_t^{(i)},t^{(i)},x^b_i\bigr)\).  The training loss is
\[
  \mathcal{L}(\theta)
  = \frac1N\sum_{i=1}^N
    \bigl\|\,v_\theta^{(i)} - v^*_t{}^{(i)}\bigr\|^2.
\]

\section{\textsc{PnP-DA} in 4D setting}
\begin{algorithm}
\caption{PnP-DA Analysis step (4D)}
\label{alg:pnp_cmf_da_4d}
\begin{algorithmic}[t]
\Require Observations $\{y_i\}_{i=t}^{t+N}$, background state $x_t^b$, model operators $\{\mathcal{M}_i\}_{i=t}^{t+N-1}$, observation operators $\{\mathcal{H}_i\}_{i=t}^{t+N}$, observation error covariances $\{\textbf{P}_i\}_{i=t}^{t+N}$, denoiser $D_\tau$, pseudo-time sequence $(\tau_n)_n$, pseudo-time learning-rate $(\gamma_n)_n$
\State Initialize: $x_t^{(0)} = x_t^b$
\For{$n = 0, 1, \ldots,$}
    \State $w^{(n)} = x_t^{(n)} - \gamma_n \nabla \left( \frac{1}{2} \sum_{i=t}^{t+N} \|y_i - \mathcal{H}_i(\mathcal{M}_{i:t}(x_t^{(n)}))\|^2_{\textbf{P}_i^{-1}} \right)$ 
    \State $\tilde{w}^{(n)} = (1 - \tau_n)z + \tau_n z^{(n)}, z \sim P_Z$ 
    \State $x_t^{(n+1)} = D_{\tau_n}(x_t^b, \tilde{w}^{(n)})$ 
\EndFor\\
\Return Analysis state $x_t^a$
\end{algorithmic}
\end{algorithm}
The \textsc{PnP‐DA} pipeline is exceptionally flexible: once the OT Bayesian flow‐matching prior has been trained, it can be dropped into either 3D‑Var or 4D‑Var style assimilation windows without retraining or architectural changes. In the 3D setting, the denoiser acts on each snapshot independently, while in 4D applications it can be interleaved with gradient steps over an entire assimilation window, seamlessly leveraging temporal correlations. The only practical requirement is that the underlying dynamical model—or its surrogate—must admit efficient Jacobian‐vector products (i.e.\ an adjoint) or be fully differentiable, so that the gradient of the observation misfit can be computed at each step. With that in place, \textsc{PnP‑DA} unifies variational assimilation and learned generative priors into a single, plug‐and‐play framework that scales naturally from static to time‑windowed DA. This will be studied in future work along with real atmopshere/weather models.

\section{Ablation Studies}
\begin{figure}[t]
\centering
\centerline{\includegraphics[width=\columnwidth]{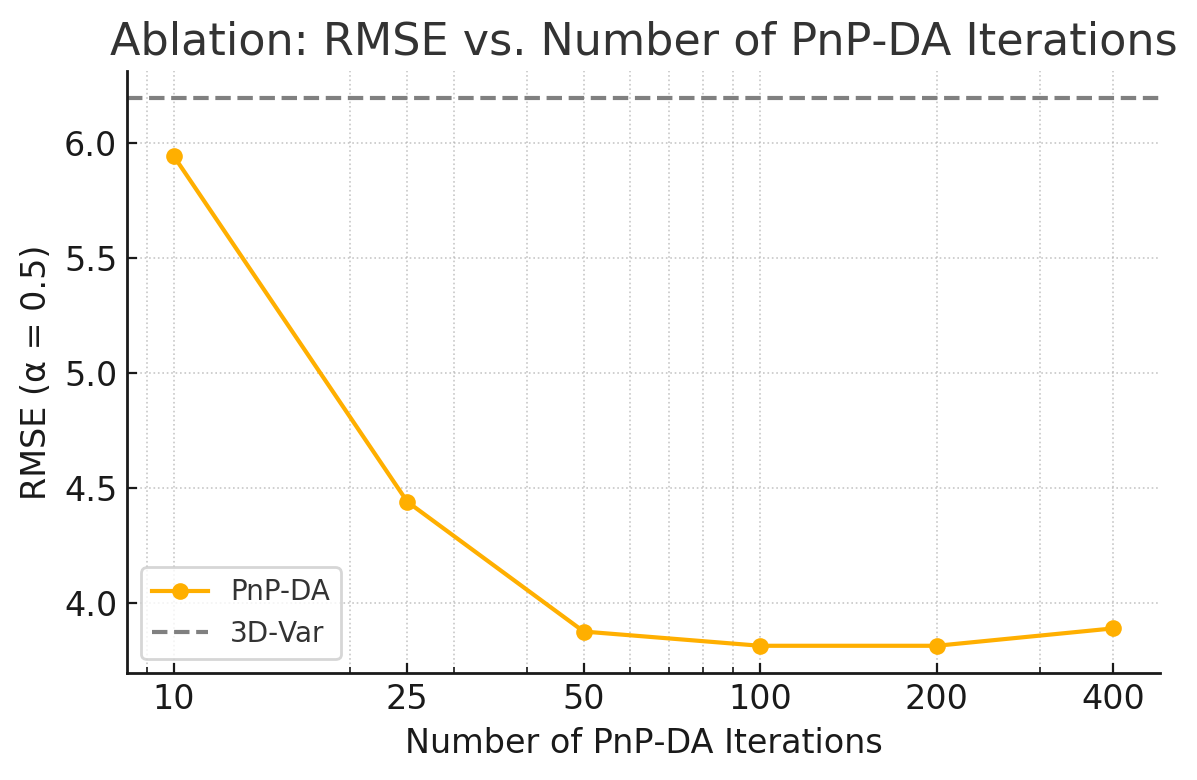}}
\caption{RMSE vs Number of \textsc{PnP-DA} Iterations}
\label{fig:ablation1}
\end{figure}

\paragraph{Number of \textsc{PnP-DA} iterations} We evaluate the sensitivity of our \textsc{PnP‑DA} method to the number of PnP iterations under the same Lorenz‑63 cyclic data assimilation setup as in the main text. Specifically, we ran 50 independent experiments for each iteration count, using $\alpha=0.5$ and all other parameters identical to the paper’s Lorenz‑63 DA experiments. The mean RMSE across these runs is plotted in Figure \ref{fig:ablation1}. As the number of iterations increases from 10 to 100, RMSE drops sharply—from about 5.94 to 3.82—before plateauing around 5000 iterations, and slightly rising at 400. The horizontal dashed line indicates the 3D‑Var baseline RMSE (6.19), demonstrating that even a small number of PnP‑DA iterations outperforms classical variational assimilation.

\begin{figure}[t]
\centering
\centerline{\includegraphics[width=\columnwidth]{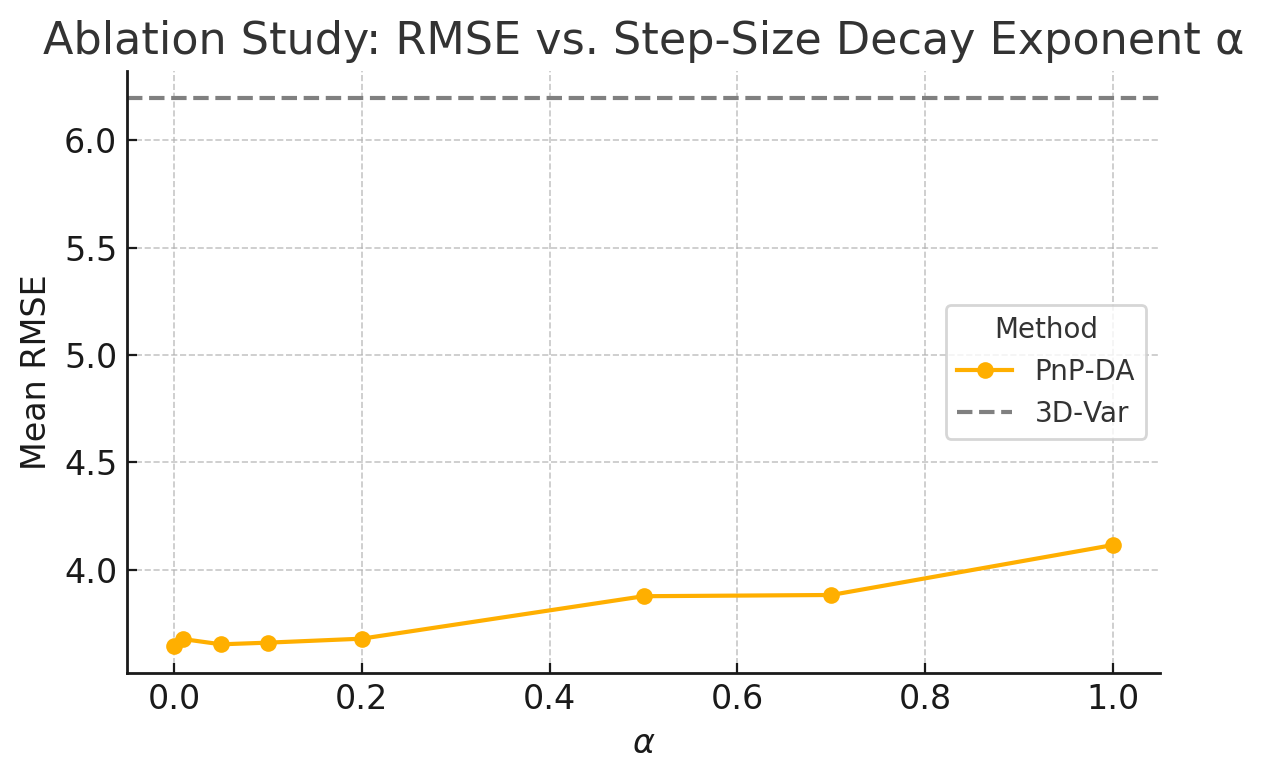}}
\caption{RMSE vs Number of \textsc{PnP-DA} Iterations}
\label{fig:ablation1}
\end{figure}

\paragraph{The choice of $\alpha$} In our implementation, $\alpha$ governs the gradient updating schedule via $\gamma_\tau = (1-\tau)^\alpha$ The ablation results (Figure 2) reveal that very small $\alpha$ values (0–0.05) minimize RMSE. This suggests that maintaining a strong influence from the flow‑matching prior across iterations leads to more accurate analyses, and that overly steep decay can underutilize the learned correction. For robustness across noise regimes and to balance convergence speed with solution quality, we adopt $\alpha=0.5$ in our main experiments, though marginal gains could be realized by tuning it toward the lower end of the tested range.

\end{document}